\documentclass{article}

\usepackage[preprint]{neurips_2026}

\usepackage[utf8]{inputenc} 
\usepackage[T1]{fontenc}    
\usepackage{hyperref}       
\usepackage{url}            
\usepackage{booktabs}       
\usepackage{amsfonts}       
\usepackage{nicefrac}       
\usepackage{microtype}      
\usepackage{xcolor}         

\usepackage{amssymb}
\usepackage{amsmath}
\usepackage{bm}
\usepackage{multirow}
\usepackage[pdftex]{graphicx}
\usepackage{float}
\usepackage{wrapfig}
\usepackage{framed}
\usepackage{tabularx}
\usepackage{makecell}
\usepackage{colortbl}
\usepackage{soul}
\usepackage{epsfig}
\usepackage{tikz}
\usepackage{pifont}
\usepackage{caption}
\usepackage{subcaption}
\usepackage{cleveref}
\usepackage{xspace}

\usepackage{enumitem}
\usepackage[flushleft]{threeparttable}

\makeatletter
\DeclareRobustCommand\onedot{\futurelet\@let@token\@onedot}
\def\@onedot{\ifx\@let@token.\else.\null\fi\xspace}

\newcommand{\MII}{\textit{MedImageInsight}\xspace}
\newcommand{\QRad}{\textit{Q}Rad\xspace}

\def\eg{\emph{e.g}\onedot}

\def\etc{\emph{etc}\onedot}

\makeatother

\title{RadFusion: Towards Threshold-Controllable Radiology Report Generation}

\author{%
Ying Jin ~ Noel C. F. Codella ~ John Corring ~ Mu Wei ~ Dinei Florencio ~ Eric Horvitz \\\\
Microsoft \\\\
\texttt{\{ying.jin, ncodella, john.corring\}@microsoft.com} \\
\texttt{\{muhsin.wei, dinei, horvitz\}@microsoft.com} \\
}

\begin{document}

\maketitle

\begin{abstract}
Automated radiology report generation is advancing rapidly in response to the shortage of radiologists, yet unlike a perception model, existing generation models offer no control over the sensitivity--specificity trade-off of their diagnostic content. Such control is essential because clinical scenarios diverge: emergency triage prioritizes sensitivity to reduce missed findings, whereas confirmatory interpretation emphasizes specificity to limit unnecessary interventions. A single fixed report can neither adapt to these scenarios nor support the ROC-based validation widely expected for regulatory clearance. We introduce RadFusion, a framework that equips report generation with threshold controllability. Our method fuses a multi-label classifier, which provides per-disease confidence scores, with a VQA-based report generator, which describes medical findings in detail; an LLM then rewrites the report so that its stated diagnoses follow the classifier's decisions at the selected threshold while staying grounded in the generator's descriptions. On MIMIC-CXR, the performance of RadFusion conforms to the classifier's ROC curve: sweeping the threshold and mapping the reports back to class labels reproduces the classifier's validated ROC performance. This conformance makes generated reports quantitatively evaluable through ROC analysis, strengthening the case for regulatory clearance, and enables operating-point selection that matches report behavior to clinical context. Moreover, combining the two model types improves diagnostic accuracy over uncontrolled generation: sensitivity increases by 6.9\% at matched specificity, and specificity by 20.7\% at matched sensitivity. These results show that RadFusion makes report generation clinically adaptable, quantitatively verifiable, and diagnostically more reliable.

\end{abstract}

\section{Introduction}
Interpreting medical images and communicating the findings in written reports is a central task in clinical care, and the growing shortage of radiologists makes its automation increasingly valuable. Clinical use, however, demands more than a fixed, plausible report: diagnostic behavior must adapt to diverse clinical scenarios and support regulatory validation. Perception models such as image classifiers meet these demands because their numerical confidence scores can be thresholded to trade off sensitivity against specificity and support ROC-based validation widely used in regulatory approval processes such as FDA clearance~\citep{food2007drug}. This threshold control is equivalent to choosing an operating point on the model's ROC curve, serving diverse clinical needs: emergency triage calls for low thresholds that prioritize sensitivity to reduce missed findings, whereas confirmatory assessments call for high thresholds that prioritize specificity to limit unnecessary interventions. However, perception models are constrained by predefined classes: a classifier may assign an 87\% probability to pneumonia, but cannot describe its anatomical location, extent, or underlying cause. Report generation models offer complementary capabilities by producing expressive, open-vocabulary text that is standard in clinical reporting practice. However, these models lack confidence estimates or threshold-based mechanisms, making them difficult to accurately assess and adjust, and ultimately leaving a gap toward trustworthy, deployable clinical AI~\citep{lin2024trustworthy}.

Radiology report generation is typically framed as an image captioning task and is advancing rapidly with recent progress in large vision--language models~\citep{llavarad,maira1,chen2024chexagent,zhou2024medversa,zhang2025libra,medpalm}. Despite the improved performance, this captioning formulation yields a single, fixed report with no mechanism to adjust the diagnostic operating point. Although controllability and confidence estimation for text generation have been studied in the general domain, existing work does not control the generated text through class-specific thresholds. For example, one line of work~\citep{dathathri2019plug,yang2021fudge} studies controllable text generation, which steers generation toward desired attributes at the token level, and another line of work~\citep{kadavath2022language,xiong2024can} focuses on the confidence-based self-assessment of candidate answers. To the best of our knowledge, threshold-controllable text generation remains a critical yet under-explored problem for clinical deployment and regulatory approval.

To fill this gap, we introduce RadFusion, a threshold-controllable radiology report generation framework that aligns the diagnostic content of generated reports with the ROC characteristics of a medical image classifier. RadFusion fuses three components: (1) a classification model that produces calibrated per-class confidence scores, (2) a VQA-based report generation model that produces detailed radiology reports and supports follow-up queries for additional details, and (3) an off-the-shelf large language model that rewrites the report so that its stated diagnoses are consistent with threshold-adjusted classification, while remaining grounded in the report generation model's findings to preserve clinical specificity. By varying the thresholds, the system produces reports spanning a range of diagnostic operating points, each corresponding to a distinct sensitivity--specificity trade-off. Moreover, since the output text is factually consistent with the classifier, the numerical confidence estimates serve as a complement to the textual report, enabling ROC-based regulatory validation and easing integration into automated systems.

RadFusion is a portable framework, generally applicable to a wide range of classifiers and report generation models. In this paper, we further study how various implementations of each component affect the final performance and identify the optimal configuration through systematic comparison. As a result, experiments on MIMIC-CXR~\citep{johnson2019mimiccxr} show that our threshold-adjusted reports exhibit the expected concave ROC behavior, with area under the curve closely tracking that of the underlying classifier. In addition, threshold control yields improvements in diagnostic accuracy relative to uncontrolled generation, including a 6.9\% increase in sensitivity at matched specificity and a 20.7\% increase in specificity at matched sensitivity. Collectively, these results demonstrate three contributions of this work:
\begin{itemize}[leftmargin=2em]
    \item \textbf{Clinical adaptability}: clinician-customizable reporting allows operating-point selection to match the clinical context.
    \item \textbf{Regulatory validation}: ROC-based evaluation of generated reports provides a viable path toward regulatory approval.
    \item \textbf{Diagnostic accuracy}: fusing complementary model types improves diagnostic accuracy over uncontrolled generation.
\end{itemize}
\section{Related Work}

\textbf{Radiology report generation.}
Early work adapted encoder--decoder architectures to radiology report generation: R2Gen~\citep{chen-emnlp-2020-r2gen} introduced memory-driven transformers, R2GenCMN~\citep{chen2022cross} added cross-modal memory, and CvT2DistilGPT2~\citep{nicolson2023improving} combined vision transformers with distilled decoders. Recent approaches leverage large vision--language models, including LLaVA-Rad~\citep{llavarad}, MAIRA~\citep{maira1,maira2}, CheXagent~\citep{chen2024chexagent}, MedVersa~\citep{zhou2024medversa}, Libra~\citep{zhang2025libra}, MedGemma~\citep{sellergren2025medgemma}, and Med-PaLM M~\citep{medpalm}. Complementary strategies, including knowledge-driven modeling~\citep{li2019knowledge_qrad,zhang2020radiology_kg}, prompt-based methods~\citep{jin2024promptmrg}, region-guided generation~\citep{rgrg}, and retrieval augmentation~\citep{sun2025factaware}, have improved factual grounding. \textit{Q}Rad~\citep{jinqrad}, a state-of-the-art model based on a pre-trained \MII~\citep{mi2} encoder, unifies the report generation and medical VQA tasks, and serves as the report generation component of our framework. All these models, however, produce a single, fixed report with no mechanism to adjust diagnostic behavior or select an operating point.

\textbf{Medical visual question answering.}
Medical VQA offers an interface for extracting clinical information via natural language queries~\citep{lau2018dataset,he2020pathvqa,liu2021slake}. Multimodal models such as Med-Flamingo~\citep{MedFlamingo23} and LLaVA-Med~\citep{llava-med} adapt vision--language architectures to medicine, while Rad-ReStruct~\citep{radrestruct} proposed hierarchical VQA for structured reporting and RaDialog~\citep{radialog} combined report generation with multi-turn dialogue. Since omissions are common in radiology reports, our framework uses medical VQA to recover omitted information from the image, supplying grounded descriptions when a low threshold flips a finding to positive.

\textbf{Controllable text generation.}
Controllable generation methods such as PPLM~\citep{dathathri2019plug}, FUDGE~\citep{yang2021fudge}, and Classifier-free Guidance~\citep{ho2022classifierfree} steer token probabilities during decoding toward desired attributes. While effective for stylistic or semantic control, these methods do not support class-specific diagnostic thresholds or operating-point selection. 

\textbf{Confidence estimation and calibration.}
Existing work on confidence estimation for LLMs includes logit-based methods using $P(\text{True})$~\citep{kadavath2022language}, verbalized confidence~\citep{xiong2024can,li2025conftuner}, sampling-based self-consistency~\citep{wang2023selfconsistency}, and the information geometry of chain-of-thought outputs~\citep{lau2025uncertainty}. Calibration then aligns the predicted confidence with empirical correctness, typically through post-hoc temperature scaling~\citep{guo2017calibration,platt1999probabilistic} or proper scoring rules during training~\citep{blasiok2023does,frohlich2024scoring}; recent task-specific methods partition the input space with uniform mass binning to approximate Bayes-optimal calibrators~\citep{manggala2025qacalibration}. For radiology report generation, \citet{wang2024trust} considered confidence during training by weighting losses with uncertainty, but offered no control mechanism at inference. Our framework builds on these foundations: per-class confidence scores are calibrated via temperature scaling so that thresholds map predictably to operating points, and logit-based $P(\text{Yes})$ extraction offers an alternative classifier implementation.
\section{Method}
\label{sec:method}

\subsection{Overview}
\label{sec:method_overview}

RadFusion enables threshold-controllable report generation by \emph{conforming} the diagnostic decisions of the generated report to those of a classification model operating at a chosen threshold, while preserving the generative model's clinical expressiveness. It is a portable framework that is applicable to a wide range of classifiers and report generation models. In this section, we first introduce the implementation that yields the optimal overall performance, followed by alternative designs for each component; the comparison among these implementations is discussed in \Cref{sec:ablation}.

\begin{figure}[t]
    \centering
    \includegraphics[width=1.0\linewidth]{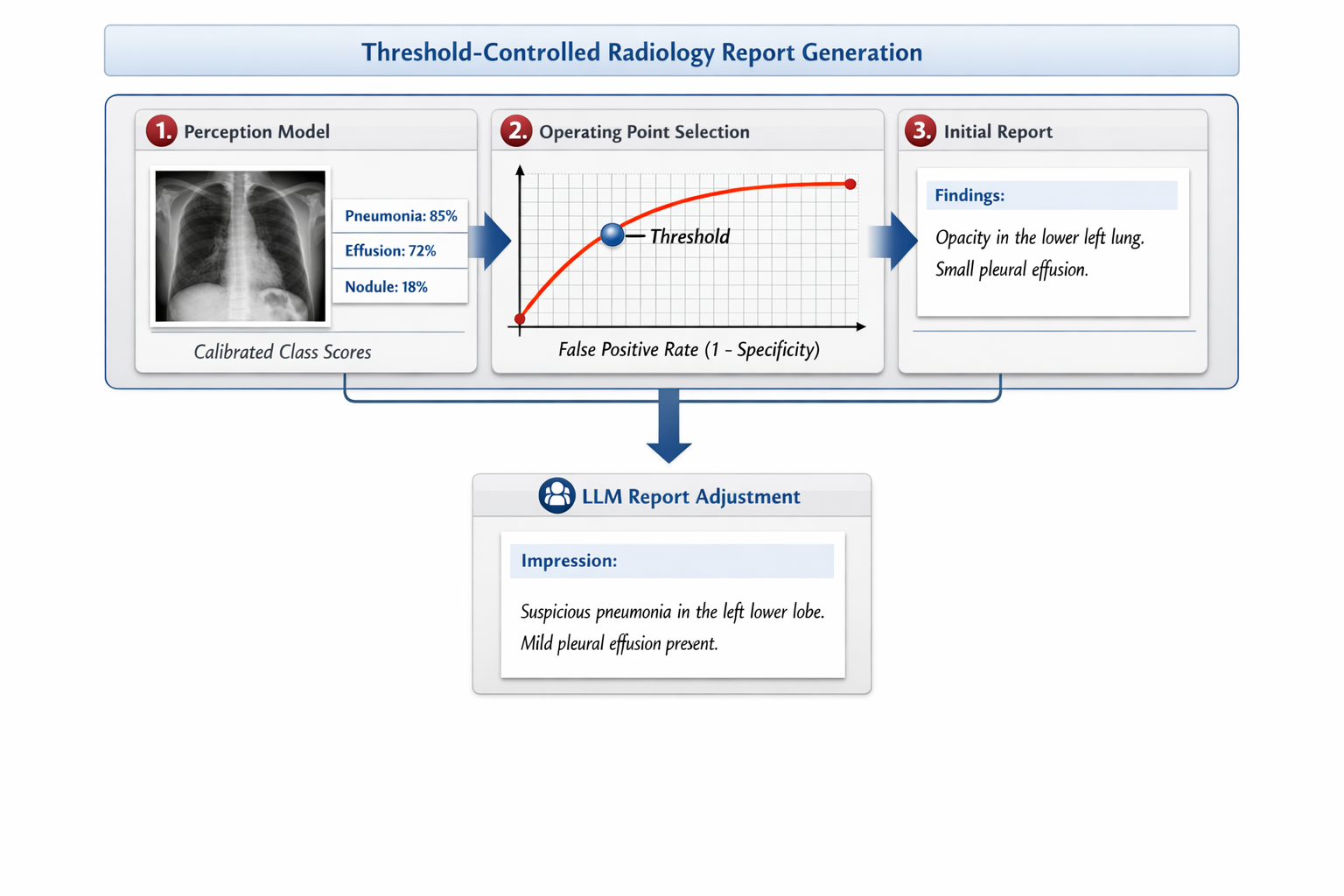}
    \caption{Overview of RadFusion, our threshold-controllable report generation framework. The classification model produces confidence scores for predefined disease categories, which are thresholded to binary decisions. The report generation model produces an initial report. The LLM rewriter combines both outputs so that the report's stated diagnoses conform to the binary classifications while preserving clinical detail. We use \QRad, a report generation model supporting follow-up questions to obtain evidence omitted from the initial report.}
    \label{fig:pipeline}
\end{figure}

As illustrated in \Cref{fig:pipeline}, RadFusion fuses three components:
\begin{enumerate}[leftmargin=2em]
    \item \textbf{Perception model} (\Cref{sec:classifier}): Given a chest X-ray image, we use a classifier to produce numerical confidence scores $\{P(C_i = 1)\}_{i=1}^{K}$ for $K$ predefined disease classes. Given a threshold $\tau$, it yields binary predictions $\hat{y}_i = \mathbf{1}[P(C_i = 1) > \tau]$, partitioning classes into positive (disease present) and negative (disease absent).
    \item \textbf{Report generation model} (\Cref{sec:qrad}): Given the same image, a report generation model produces a free-text radiology report $r$ describing both positive and negative findings. Compared to the classifier's binary labels, this open-vocabulary report contains richer information, including the anatomical location, severity, progression, and clinical context of each finding.
    \item \textbf{LLM rewriter} (\Cref{sec:rewriter}): An off-the-shelf large language model that combines the classifier's binary decisions $\hat{y}$ with the report's detailed descriptions $r$, rewriting the report so that its stated diagnoses align with the thresholded classes while remaining grounded in the clinical attributes from the original report.
\end{enumerate}

By sweeping a threshold $\tau$ from 0 to 1 per class, RadFusion produces a family of reports spanning the full sensitivity--specificity spectrum. For evaluation, we assess the diagnostic quality of these reports across operating points with a closed-loop protocol detailed in \Cref{sec:threshold_eval}.

\subsection{Perception Model: MedImageInsight Classifier}
\label{sec:classifier}

Given a chest X-ray image, the classification model produces numerical confidence scores for 14 disease classes defined by CheXpert~\citep{irvin2019chexpert}, such as Enlarged Cardiomediastinum, Cardiomegaly, and Lung Opacity. We study \textit{three different} classification approaches. The baseline approach is based on \MII (MI2)~\citep{mi2}, a medical image-text foundation model with a DaViT~\citep{davit} image encoder (360M parameters) and a language encoder (252M parameters), pre-trained via contrastive learning~\citep{unicl} on over 3.5 million medical image-text pairs across 12 imaging domains. A classification model is fine-tuned and calibrated on MIMIC-CXR~\citep{johnson2019mimiccxr} as described below. Its performance evaluation is discussed in \Cref{sec:experiments}.

\textbf{Classifier Approach 1: Baseline.}
To adapt MI2 to the CheXpert classes, we fine-tune it with the Image-Text-Class Hybrid Contrastive loss~\citep{jindaug}. This hybrid loss enables the model to jointly leverage image-class and image-text supervision to attain superior classification performance. The structured class labels ground the model on the disease classes of interest, while the free-text reports provide richer, fine-grained supervision that prevents the feature space from collapsing onto the coarse class labels alone. The loss function is defined as:
\begin{equation}
    \mathcal{L} = \mathcal{L}_{\text{image-text}} + \lambda \mathcal{L}_{\text{image-class}},
    \label{eq:hybrid_loss}
\end{equation}
where $\mathcal{L}_{\text{image-text}}$ is the image-text contrastive loss based on softmax cross-entropy, and $\mathcal{L}_{\text{image-class}}$ is the image-class contrastive loss based on sigmoid cross-entropy (the sigmoid accommodates the multi-label setup). The embeddings for the classes are based on template prompts (e.g., ``a chest X-ray showing [CLASS NAME]''). We use class labels from VisualCheXbert~\citep{jain2021visualchexbert}. At inference, the class probabilities are computed as:
\begin{equation}
    P(C_i = 1 \mid I) = \text{sigmoid}[(\text{cosine\_similarity}(f_I(I),\; f_T(t_i)) / \tau_{\text{temp}})],
    \label{eq:cls_score}
\end{equation}
where $f_I$, $f_T$ are the image and text encoders, $t_i$ is the class prompt, and $\tau_{\text{temp}}$ is a learnable temperature. Since the class prompts are fixed, we precompute $f_T(t_i)$ as the weights of a single linear classification head, removing the text encoder at inference time.

The confidence scores are calibrated by temperature scaling~\citep{guo2017calibration}, which involves learning a scalar $T$ that minimizes the Expected Calibration Error on a validation set. This lightweight post-hoc correction improves calibration without altering discriminative performance (AUC is unchanged).

\textbf{Classifier Approach 2: Linear probing of QRad's encoder.}
An alternative implementation of the classifier trains a linear classification head on top of \QRad's frozen image encoder, i.e., a linear probe. Since \QRad's encoder is also initialized from MI2, this alternative differs from the MI2 classifier only by an additional generative fine-tuning step; moreover, at test time the classifier and the report generation model share the same encoder, reducing computational cost. The linear probe boosts performance on hard-to-detect classes (\eg Pneumothorax) but slightly underperforms on easy-to-detect ones (\eg Support Devices), with overall ROC-AUC comparable to the MI2 classifier, as shown in \Cref{sec:ablation}.

\textbf{Classifier Approach 3: Single-token probability from QRad.}
QRad, being a hybrid report generation and medical VQA model, is trained with binary classification questions answered by single-token yes/no responses (e.g., ``Is this image classified as [CLASS]? (yes/no)''). We extract the softmax probability over the \texttt{[Yes]} and \texttt{[No]} logits~\citep{kadavath2022language} as the class confidence. This alternative has the advantage of using a single model for both numerical confidence and textual report generation, but it underperforms a standalone classifier on most classes.

\subsection{Report Generation Model: QRad Auto-VQA}
\label{sec:qrad}

Given a chest X-ray image, a report generation model produces a free-text radiology report that describes the patient's findings in clinical language. The report covers both positive findings (e.g., ``mild bibasilar atelectasis'') and pertinent negatives (e.g., ``no pneumothorax''), along with attributes that binary labels cannot express, such as the \emph{location} of a finding (right lower lobe vs.\ bilateral), its \emph{severity} (mild, moderate, severe), \emph{progression} relative to prior studies (stable, worsening, improved), and \emph{clinical context} (e.g., post-surgical changes). Omission of some negative findings is common in radiology reports.

Conventional report generation models follow a direct image-to-text mapping $Y = f(I)$, producing a monolithic report with no mechanism to query additional details that may have been omitted. We use \QRad~\citep{jinqrad}, which builds on the MI2 vision encoder and reframes report generation as a self-directed Visual Question Answering (Auto-VQA) process:
\begin{equation}
    Q = f_Q(I), \quad Y = f_A(I, Q),
\end{equation}
where a Question Generator $f_Q$ predicts a sequence of clinically relevant questions conditioned on the image, and an Answer Generator $f_A$ answers each question to produce one sentence of the report. The final report is the concatenation of all answers. This Auto-VQA reframing allows RadFusion to query disease categories omitted in the initial report by composing questions and sending them to the Answer Generator $f_A$: when a low threshold flips a class to positive, a follow-up query returns a grounded description of the finding with its location, severity, and context, rather than leaving the rewriter to hallucinate such details. The initial report and all follow-up answers together form the evidence pool for rewriting.

\textbf{Alternative approaches:}
RadFusion is widely applicable to report generation models, such as UniRG-CXR~\citep{liu2026scaling}, which leverages reinforcement fine-tuning. For models without a query interface, a separate VQA model may be required to recover omitted information. Because \QRad serves both roles with a single model, we adopt \QRad in this study as a simple yet effective implementation.

\subsection{LLM Rewriting}
\label{sec:rewriter}

The LLM rewriter combines the outputs of the two preceding components: the binary predictions $\{\hat{y}_i\}_{i=1}^{K}$ from the thresholded classifier, and the evidence pool $\mathcal{E}$ consisting of the initial report $r$ and optional follow-up answers. We use an off-the-shelf LLM such as GPT-5 \citep{gpt5} to rewrite the report\footnote{We use an on-prem deployment to satisfy dataset usage agreements.} because modern LLMs have demonstrated strong instruction-following capabilities, avoiding the need to train a dedicated model for this task.

The rewriting instruction for the LLM plays a critical role in ensuring that its stated diagnoses are consistent with $\hat{y}_i$ while remaining grounded in the clinical attributes (location, severity, progression, \etc) provided by $\mathcal{E}$. Therefore, we structure the instruction to explicitly define the desired behavior of the LLM, including how to handle positive and negative classes, how to incorporate evidence from $\mathcal{E}$, and how to maintain non-class content. Specifically:

\begin{itemize}[leftmargin=2em]
    \item For each positive class ($\hat{y}_i = 1$): if the finding is already present in the report, it is kept unchanged; if absent or contradicted, the LLM adds or corrects it using detail from $\mathcal{E}$.
    \item For each negative class ($\hat{y}_i = 0$): if the finding is mentioned as present, the LLM removes or negates it; if already absent, no change is needed.
    \item Non-class content (e.g., imaging quality, support devices not in the class set) is preserved.
\end{itemize}

An LLM capable of this task must have medical domain knowledge to reason over the disease class names and the input report. To help establish the correspondence between classes and their varied natural language manifestations in the report, we provide examples per class as part of the instruction. For example, we show that the class ``Pleural Effusion'' may appear in a report as ``small bilateral pleural effusions'', ``fluid in the costophrenic angles'', or ``blunting of the costophrenic recesses''. The full instruction is provided in \Cref{sec:rewriting_prompt}.

\textbf{Alternative approaches: other LLMs as the rewriter.}
RadFusion can employ any capable LLM as the rewriter, and satisfactory performance depends on the model's medical knowledge, reasoning, and instruction-following capabilities. In \Cref{sec:ablation}, we compare alternatives spanning models of varied sizes, including GPT 5.4, GPT 5.4-mini, and DeepSeek V4 Pro, as well as varied levels of reasoning effort.

\subsection{Extension to Three-Dimensional Threshold Control for Clinical Deployment}
\label{sec:3d_threshold}
RadFusion's rewriting-based design extends naturally beyond a single binary threshold. In clinical practice, radiologists rarely operate on a strict positive/negative dichotomy: borderline findings are reported with explicit hedging (e.g., ``cannot exclude early consolidation''), and the appropriate response to a finding depends not only on diagnostic confidence but also on its time sensitivity. To support such graded decision-making, we extend the threshold space to three dimensions $(T_a, T_b, T_c)$ per class, which requires only modified rewriting instructions rather than any model retraining. The first two dimensions ($T_a \leq T_b$) partition the confidence spectrum into three zones:
\begin{itemize}[leftmargin=2em]
    \item $P(C_i = 1 \mid I) < T_a$: \emph{negative zone}, where the report denies the finding.
    \item $P(C_i = 1 \mid I) > T_b$: \emph{positive zone}, where the report asserts the finding.
    \item $T_a \leq P(C_i = 1 \mid I) \leq T_b$: \emph{semi-positive zone}, where the report describes the finding while explicitly flagging uncertainty and recommending further evaluation.
\end{itemize}
The third dimension, $T_c$, thresholds a per-class \emph{urgency} score, motivated by the fact that clinical urgency does not always align with imaging severity (e.g., an early pneumothorax may demand immediate treatment); the score can be derived from clinical priors, patient context, or an auxiliary classifier, and findings exceeding $T_c$ are emphasized as time-sensitive and actionable, independent of the $T_a$/$T_b$ axis. The system presented in \Cref{sec:method_overview,sec:rewriter} is the special case with $T_a = T_b$ and $T_c$ omitted, which we adopt for direct ROC-based comparison against the underlying classifier. This extension illustrates the extendability of the design: richer reporting policies, such as asserting findings conservatively (high $T_b$) while escalating asserted ones aggressively (low $T_c$), are implemented by adjusting thresholds and instructions alone. We view clinician-tunable operating-point interfaces as a promising direction for future work.
\section{Experiments}
\label{sec:experiments}

\subsection{Dataset and Setup}

We conduct experiments on MIMIC-CXR~\citep{johnson2019mimiccxr}, one of the largest radiology report datasets containing 227,835 studies with 377,110 chest X-ray images. Following standard practice~\citep{llavarad,maira1}, we use the Findings section as the generation target. All evaluations use the official test split (2,347 studies).

\subsection{Evaluation Methods}
\label{sec:threshold_eval}

To assess the threshold controllability of RadFusion, we implement a closed-loop evaluation protocol that runs from the classifier's thresholded decisions, through the rewritten report, and back to class labels, which are then compared against the original decisions for alignment. Given a threshold $\tau$, the classifier's confidence scores are binarized into positive and negative class lists, and the LLM rewrites \QRad's initial report to conform to these classifications, as described in \Cref{sec:rewriter}. To evaluate, we convert the rewritten report back to binary class labels using GPT-5, which we found more reliable than VisualCheXbert~\citep{jain2021visualchexbert}, a BERT-based text classification model commonly used to extract disease classes from radiology reports. Comparing the extracted labels against the ground truth yields the true positive rate (sensitivity) and false positive rate ($1 - \text{specificity}$) of the rewritten reports at threshold $\tau$.

\begin{figure}[t]
    \centering
    \includegraphics[width=\linewidth]{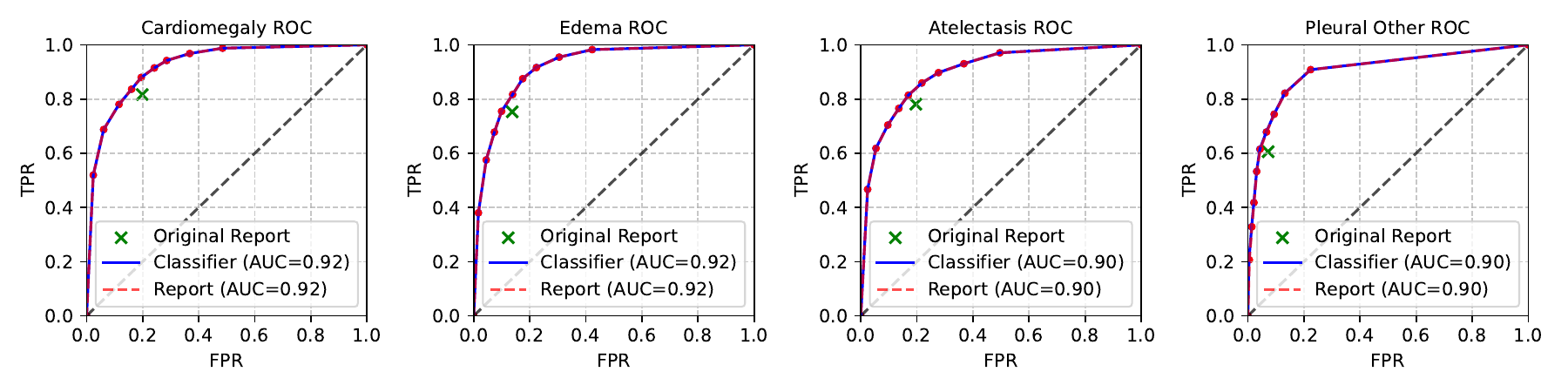}
    \caption{Threshold-controllable report generation on four representative classes. \textbf{Cardiomegaly}: common, both models perform strongly. \textbf{Edema}: threshold tuning is clinically valuable since mild cases often need no treatment. \textbf{Atelectasis}: challenging, requires fine visual details. \textbf{Pleural Other}: rare (1.3\% prevalence). Across all four, the report ROC (red) closely tracks the classifier (blue), and the non-controlled report (green cross) falls below the curves.}
    \label{fig:threshold_ctrl_4_clases}
\end{figure}

We sweep $\tau$ from 0.0 to 1.0 with a step size of 0.1, yielding 11 operating points.\footnote{We find 0.1 is a reasonable step size for the threshold. Smaller step sizes produce smoother curves, but increase LLM inference cost linearly.} The same $\tau$ is applied across all 13 disease classes (``No Finding'' is excluded because it semantically overlaps with the absence of other classes). Connecting these operating points traces the ROC curve of the threshold-controlled reports. The ROC curve plots TPR against FPR across thresholds, and each of its points corresponds to a specific sensitivity--specificity trade-off, making ROC the natural evaluation for threshold-controllable systems. It is also the metric widely expected by regulatory approval processes for medical devices. The area under the curve (AUC) summarizes the discriminative performance across all operating points on the ROC curve. We compare both the ROC curves and their AUC among three systems (\Cref{fig:threshold_ctrl_4_clases,fig:threshold_ctrl_full_supp}):
\begin{itemize}[leftmargin=2em]
    \item \textbf{The classifier (blue curves)}: ROC computed directly from the classifier's continuous confidence scores, the upper bound we aim to conform to.
    \item \textbf{Threshold-controlled reports (red curves)}: ROC computed by converting each threshold-adjusted report back to class labels. Each point corresponds to one operating point specified by a threshold.
    \item \textbf{Non-controlled reports (green cross)}: The single operating point for the generated report without threshold adjustments.
\end{itemize}

\subsection{Results and Discussion}

Overall, the results demonstrate that RadFusion delivers effective threshold control: the threshold-controlled reports closely conform to the classifier's ROC curves, improve diagnostic accuracy over non-controlled generation, and expose a tunable sensitivity--specificity trade-off that serves diverse clinical scenarios. We highlight three key findings:

\textbf{Conformance to classifier ROC.} \Cref{fig:threshold_ctrl_4_clases} highlights four classes spanning a range of difficulty and prevalence: Cardiomegaly (common), Edema (threshold-sensitive), Atelectasis (hard), and Pleural Other (rare, 1.3\%). Across all four, the report ROC closely tracks the classifier ROC with comparable AUC, and the concave shape confirms a monotonic sensitivity--specificity trade-off; the full 13-class evaluation (\Cref{fig:threshold_ctrl_full_supp}) shows the same pattern. This close overlap validates that the threshold effectively trades off sensitivity and specificity, and that the rewritten reports conform to the classifier's decisions across the operating range.

\textbf{Improved factual correctness.} The ROC curves of threshold-controlled reports are \emph{above} the non-controlled reports (green crosses): at matched specificity, sensitivity improves by 6.9\%; at matched sensitivity, specificity improves by 20.7\%. By combining the two model types, threshold control transfers the classifier's superior discriminative performance to the generated reports.

\textbf{Clinical utility.} Low thresholds prioritize sensitivity (screening/triage); high thresholds prioritize specificity (confirmatory assessment). Our framework makes this trade-off explicit and tunable, replacing the fixed, opaque operating point of conventional models and offering a path toward ROC-based regulatory evaluation of generated reports.

\section{Ablation Studies}
\label{sec:ablation}

\subsection{Alternative Classifier Implementations}
\label{sec:classifier_abl}

Here we compare the three classifier implementations detailed in \Cref{sec:classifier}: a standalone classifier fine-tuned from MI2 (our default), a linear probe of \QRad's frozen encoder (sharing the encoder with report generation), and single-token yes/no probabilities extracted from \QRad's VQA answers (a single model for classification and report generation). All three implementations use the same \MII pre-trained image encoder, and \QRad further fine-tunes the encoder for report generation.

\Cref{tab:roc_auc} reports their per-class AUC-ROC, and their ROC curves are provided in \Cref{sec:roc_classifiers}. The MI2 classifier and the \QRad linear probe achieve comparable overall AUC, each stronger on a different subset of classes, while the \QRad token-logit variant trails on most classes. Notably, the \QRad Linear Prob implementation significantly improves on hard classes like Pneumothorax, possibly because the report generation objective enhances the encoder's visual representation to better capture subtle visual cues. The MI2 FT classifier is stronger on visually obvious classes like Support Devices, possibly due to its pretraining on a large corpus of radiology images. RadFusion uses the MI2 FT classifier by default to demonstrate a common use case where separate classifiers and report generators are combined for better utility than either alone.

\begin{table}[t]
\centering
\setlength{\tabcolsep}{2pt}
\caption{AUC-ROC of the three classifier implementations. MI2 FT: the fine-tuned classifier based on \MII; \QRad Linear Prob: fitting a linear classification head on \QRad's frozen encoder; \QRad Token Logit: single-token yes/no probabilities using \QRad's VQA capability. Results show AUC-ROC across the 13 disease classes, with AVG their macro average.}
\renewcommand{\arraystretch}{1.1}
\begin{tabular}{lccccccccccccc|c}
\hline
\textbf{AUC-ROC} & \textbf{Enl.} & \textbf{Car.} & \textbf{L.O.} & \textbf{L.L.} & \textbf{Ede.} & \textbf{Con.} & \textbf{Pmn.} & \textbf{Ate.} & \textbf{Pmt.} & \textbf{P.E.} & \textbf{P.O.} & \textbf{Fra.} & \textbf{S.D.} & \textbf{AVG} \\
\hline
MI2 FT           & \textbf{.93} & \textbf{.92} & \textbf{.92} & .85 & \textbf{.92} & \textbf{.91} & .88 & .90 & .88 & .90 & \textbf{.90} & \textbf{.89} & \textbf{.95} & .90 \\
\QRad Linear Prob & .92 & .91 & \textbf{.92} & \textbf{.89} & .91 & \textbf{.91} & \textbf{.89} & \textbf{.91} & \textbf{.95} & \textbf{.92} & \textbf{.90} & .87 & .91 & \textbf{.91} \\
\QRad Token Logit & .82 & .81 & .83 & .73 & .78 & .78 & .71 & .74 & .68 & .78 & .58 & .56 & .65 & .73\\
\hline
\end{tabular}
\label{tab:roc_auc}
\end{table}

\subsection{Alternative LLM Rewriters}
\label{sec:llm_abl}

\textbf{Model size and rewriting instruction.} We compare LLM rewriters across model sizes (GPT-5, GPT-5.4, and GPT-5.4-mini) and rewriting instructions, with and without the per-class example text segments (\Cref{sec:rewriter}); by default, the instruction includes the examples. \Cref{tab:llm_abl} reports the AUC-ROC of the threshold-controlled reports for each configuration. Removing the examples degrades performance: for GPT-5, the AUC-ROC on Fracture (Fra.) drops from .89 to .73 and the macro average from .90 to .88. This confirms that grounding the class--text correspondence with concrete examples is important for faithful rewriting, consistent with our analysis in \Cref{sec:rewriter}. We further observe that a newer model generation does not necessarily outperform its predecessor under the same prompt: GPT-5.4 does not exceed GPT-5. This suggests that different models may require specifically optimized instructions for the optimal performance (the instruction was tuned on GPT-5).

\begin{table}[t]
\centering
\setlength{\tabcolsep}{2pt}
\caption{Comparison of LLM rewriter implementations across model sizes and rewriting instructions. By default, the instruction includes the per-class example text segments; ``w/o examples'' removes them. The table shows AUC-ROC of the threshold-controlled reports across the 13 disease classes, with AVG their macro average. Within each model group (separated by midrules), the higher AVG is in bold.}
\renewcommand{\arraystretch}{1.2}
\resizebox{\textwidth}{!}{%
\begin{tabular}{llccccccccccccc|c}
\hline
\textbf{Model} & \textbf{Reasoning} & \textbf{Enl.} & \textbf{Car.} & \textbf{L.O.} & \textbf{L.L.} & \textbf{Ede.} & \textbf{Con.} & \textbf{Pmn.} & \textbf{Ate.} & \textbf{Pmt.} & \textbf{P.E.} & \textbf{P.O.} & \textbf{Fra.} & \textbf{S.D.} & \textbf{AVG} \\
\hline
GPT-5           & default & .93 & .92 & .92 & .84 & .92 & .91 & .88 & .90 & .88 & .90 & .90 & .89 & .93 & \textbf{.90}\\
GPT-5 w/o examples & default & .88 & .92 & .91 & .84 & .91 & .91 & .88 & .90 & .88 & .90 & .90 & .73 & .89 & .88 \\
\midrule
GPT-5.4         & medium  & .93 & .92 & .92 & .75 & .92 & .91 & .88 & .90 & .88 & .90 & .90 & .89 & .94 & .90 \\
\midrule
GPT-5.4-mini    & medium  & .93 & .92 & .92 & .82 & .90 & .90 & .88 & .90 & .84 & .90 & .90 & .88 & .93 & .89 \\
\hline
\end{tabular}%
}
\label{tab:llm_abl}
\end{table}

\textbf{Model provider and reasoning effort.} \Cref{tab:llm_abl_linprob} extends the comparison across model providers (OpenAI and DeepSeek) and reasoning efforts, using the \QRad Linear Prob classifier as the perception model for additional context. The results exhibit a consistent pattern with \Cref{tab:llm_abl}, and further show that reasoning effort is vital, especially for smaller models. Notably, both GPT-5.4 and GPT-5.4-mini experience significant degradation on Lung Lesion (L.L.), but raising the reasoning effort to medium improves GPT-5.4-mini from .58 to .81 on this class.

\begin{table}[t]
\centering
\setlength{\tabcolsep}{2pt}
\caption{Comparison of LLM rewriter implementations across model providers (OpenAI and DeepSeek) and reasoning efforts, using the \QRad Linear Prob classifier as the perception model for additional context. The pattern is consistent with \Cref{tab:llm_abl}, and reasoning effort is vital especially for smaller models. The table shows AUC-ROC of the threshold-controlled reports across the 13 disease classes, with AVG their macro average. Within each model group (separated by midrules), the higher AVG is in bold. DS-V4-Pro denotes DeepSeek V4 Pro \citep{dsv4}.}
\renewcommand{\arraystretch}{1.2}
\resizebox{\textwidth}{!}{%
\begin{tabular}{llccccccccccccc|c}
\hline
\textbf{Model} & \textbf{Reasoning} & \textbf{Enl.} & \textbf{Car.} & \textbf{L.O.} & \textbf{L.L.} & \textbf{Ede.} & \textbf{Con.} & \textbf{Pmn.} & \textbf{Ate.} & \textbf{Pmt.} & \textbf{P.E.} & \textbf{P.O.} & \textbf{Fra.} & \textbf{S.D.} & \textbf{AVG} \\
\hline
GPT-5           & default & .92 & .91 & .91 & .85 & .91 & .90 & .88 & .90 & .94 & .91 & .89 & .86 & .89 & .90 \\
\midrule
GPT-5.4         & default & .92 & .91 & .91 & .73 & .91 & .90 & .88 & .90 & .94 & .91 & .87 & .84 & .85 & .88 \\
GPT-5.4         & medium  & .92 & .91 & .91 & .73 & .91 & .90 & .89 & .90 & .94 & .91 & .89 & .86 & .88 & \textbf{.89} \\
\midrule
GPT-5.4-mini    & default & .86 & .89 & .83 & .58 & .77 & .63 & .67 & .85 & .65 & .85 & .85 & .56 & .85 & .76 \\
GPT-5.4-mini    & medium  & .92 & .91 & .91 & .81 & .90 & .90 & .87 & .90 & .88 & .90 & .89 & .85 & .88 & \textbf{.89} \\
\midrule
DS-V4-Pro       & default & .91 & .91 & .89 & .77 & .89 & .89 & .85 & .89 & .87 & .89 & .87 & .85 & .85 & .87 \\
\hline
\end{tabular}%
}
\label{tab:llm_abl_linprob}
\end{table}

\section{Conclusion}

We introduced RadFusion, a threshold-controllable radiology report generation framework that fuses a classifier, a VQA-based report generator, and an LLM rewriter. By conforming the report's diagnostic content to the classifier's threshold-adjusted predictions, RadFusion produces reports whose performance tracks the classifier's ROC curve while preserving rich clinical detail, so that a single threshold selects the operating point along the sensitivity--specificity spectrum. On MIMIC-CXR, this conformance improves sensitivity by 6.9\% at matched specificity and specificity by 20.7\% at matched sensitivity over non-controlled generation. To our knowledge, this is the first framework to achieve threshold control in report generation, making generated reports clinically adaptable, quantitatively evaluable, and diagnostically more accurate, and offering a practical path toward regulatory validation and clinician-customizable AI-assisted radiology.

\newpage
\section{Broader Impact}

RadFusion can improve the safety and clinical utility of AI-assisted radiology. By making the sensitivity--specificity trade-off explicit and tunable, it lets clinicians match model behavior to a clinical context (high sensitivity for screening, high specificity for confirmatory assessment) rather than relying on a fixed, opaque operating point, enabling responsible deployment across settings with diverse risk tolerances. Because each report is paired with the classifier's per-class confidence scores, the framework also supports the ROC-based evaluations widely expected in regulatory pathways such as FDA device authorization~\cite{food2007drug} and eases integration into rule-based clinical decision systems. Beyond radiology, conforming generative text to a perception model's structured decisions applies to any domain where generated text must make decisions with tunable risk tolerance, such as autonomous systems, content moderation, and legal compliance.

\textbf{Limitations and risks.} RadFusion relies on the quality of all three components: classifier errors propagate into the rewritten report, the report generator may omit findings, and the LLM rewriter may introduce subtle linguistic artifacts. The rewriting instruction is template-based and tuned for chest X-ray findings; our ablations show that its effectiveness varies across LLMs, so it would require re-tuning for a different rewriter, imaging modality, or clinical domain. Threshold control also operates over a predefined set of disease classes, and findings outside this set are not subject to threshold adjustment. Finally, our evaluation relies on automatic label extraction from reports, which may itself introduce error. We emphasize that our system is intended to assist, not replace, radiologist review.

{
    \bibliographystyle{ieeenat_fullname}
    \bibliography{refs}

@article{gpt5,
  title={Openai gpt-5 system card},
  author={Singh, Aaditya and Fry, Adam and Perelman, Adam and Tart, Adam and Ganesh, Adi and El-Kishky, Ahmed and McLaughlin, Aidan and Low, Aiden and Ostrow, AJ and Ananthram, Akhila and others},
  journal={arXiv preprint arXiv:2601.03267},
  year={2025}
}

@inproceedings{
manggala2025qacalibration,
title={{QA}-Calibration of Language Model Confidence Scores},
author={Putra Manggala and Atalanti A. Mastakouri and Elke Kirschbaum and Shiva Kasiviswanathan and Aaditya Ramdas},
booktitle={The Thirteenth International Conference on Learning Representations},
year={2025},
url={https://openreview.net/forum?id=D2hhkU5O48}
}

@inproceedings{
lau2025uncertainty,
title={Uncertainty Quantification for {MLLM}s},
author={Gregory Kang Ruey Lau and Hieu Dao and Bryan Kian Hsiang Low},
booktitle={ICLR Workshop: Quantify Uncertainty and Hallucination in Foundation Models: The Next Frontier in Reliable AI},
year={2025},
url={https://openreview.net/forum?id=vCqtd8ksh3}
}

@article{chen2022cross,
  title   = {Cross-modal memory networks for radiology report generation},
  author  = {Chen, Zhihong and Shen, Yaling and Song, Yan and Wan, Xiang},
  journal = {arXiv preprint arXiv:2204.13258},
  year    = {2022}
}

@inproceedings{jain2021visualchexbert,
  title     = {VisualCheXbert: addressing the discrepancy between radiology report labels and image labels},
  author    = {Jain, Saahil and Smit, Akshay and Truong, Steven QH and Nguyen, Chanh DT and Huynh, Minh-Thanh and Jain, Mudit and Young, Victoria A and Ng, Andrew Y and Lungren, Matthew P and Rajpurkar, Pranav},
  booktitle = {Proceedings of the Conference on Health, Inference, and Learning},
  pages     = {105--115},
  year      = {2021}
}

@article{johnson2019mimiccxr,
  title     = {MIMIC-CXR, a de-identified publicly available database of chest radiographs with free-text reports},
  author    = {Johnson, Alistair EW and Pollard, Tom J and Berkowitz, Seth J and Greenbaum, Nathaniel R and Lungren, Matthew P and Deng, Chih-ying and Mark, Roger G and Horng, Steven},
  journal   = {Scientific data},
  volume    = {6},
  number    = {1},
  pages     = {317},
  year      = {2019},
  publisher = {Nature Publishing Group UK London}
}

@article{li2019knowledge_qrad,
  title     = {Knowledge driven temporal activity localization},
  author    = {Li, Changlin and Li, Zhihui and Ge, Zongyuan and Li, Mingjie},
  journal   = {Journal of Visual Communication and Image Representation},
  volume    = {64},
  pages     = {102628},
  year      = {2019},
  publisher = {Elsevier}
}

@article{zhou2024medversa,
  title={MedVersa: A Generalist Foundation Model for Medical Image Interpretation},
  author={Zhou, Hong-Yu and Acosta, Juli{\'a}n Nicol{\'a}s and Adithan, Subathra and Datta, Suvrankar and Topol, Eric J and Rajpurkar, Pranav},
  journal={arXiv preprint arXiv:2405.07988},
  year={2024}
}

@inproceedings{zhang2025libra,
  title={Libra: Leveraging temporal images for biomedical radiology analysis},
  author={Zhang, Xi and Meng, Zaiqiao and Lever, Jake and Ho, Edmond SL},
  booktitle={Findings of the Association for Computational Linguistics: ACL 2025},
  pages={17275--17303},
  year={2025}
}

@inproceedings{chen-emnlp-2020-r2gen,
  title     = {Generating Radiology Reports via Memory-driven Transformer},
  author    = {Chen, Zhihong and
               Song, Yan  and
               Chang, Tsung-Hui and
               Wan, Xiang},
  booktitle = {Proceedings of the 2020 Conference on Empirical Methods in Natural Language Processing},
  month     = nov,
  year      = {2020}
}

@article{nicolson2023improving,
  title={Improving chest X-ray report generation by leveraging warm starting},
  author={Nicolson, Aaron and Dowling, Jason and Koopman, Bevan},
  journal={Artificial intelligence in medicine},
  volume={144},
  pages={102633},
  year={2023},
  publisher={Elsevier}
}

@article{chen2024chexagent,
  title   = {Chexagent: Towards a foundation model for chest x-ray interpretation},
  author  = {Chen, Zhihong and Varma, Maya and Delbrouck, Jean-Benoit and Paschali, Magdalini and Blankemeier, Louis and Van Veen, Dave and Valanarasu, Jeya Maria Jose and Youssef, Alaa and Cohen, Joseph Paul and Reis, Eduardo Pontes and others},
  journal = {arXiv preprint arXiv:2401.12208},
  year    = {2024}
}

@article{food2007drug,
  title   = {Drug Administration. Statistical guidance on reporting results from studies evaluating diagnostic tests-guidance for industry and FDA staff},
  author  = {Food, US},
  journal = {Food and Drug Administration, Center for Devices and Radiological Health Diagnostic Devices Branch, Division of Biostatistics, Office of Surveillance and Biometrics},
  year    = {2007}
}

@inproceedings{irvin2019chexpert,
  title     = {Chexpert: A large chest radiograph dataset with uncertainty labels and expert comparison},
  author    = {Irvin, Jeremy and Rajpurkar, Pranav and Ko, Michael and Yu, Yifan and Ciurea-Ilcus, Silviana and Chute, Chris and Marklund, Henrik and Haghgoo, Behzad and Ball, Robyn and Shpanskaya, Katie and others},
  booktitle = {Proceedings of the AAAI conference on artificial intelligence},
  volume    = {33},
  number    = {01},
  pages     = {590--597},
  year      = {2019}
}

@inproceedings{jin2024promptmrg,
  title     = {Promptmrg: Diagnosis-driven prompts for medical report generation},
  author    = {Jin, Haibo and Che, Haoxuan and Lin, Yi and Chen, Hao},
  booktitle = {Proceedings of the AAAI Conference on Artificial Intelligence},
  volume    = {38},
  number    = {3},
  pages     = {2607--2615},
  year      = {2024}
}

@article{kadavath2022language,
  title   = {Language models (mostly) know what they know},
  author  = {Kadavath, Saurav and Conerly, Tom and Askell, Amanda and Henighan, Tom and Drain, Dawn and Perez, Ethan and Schiefer, Nicholas and Hatfield-Dodds, Zac and DasSarma, Nova and Tran-Johnson, Eli and others},
  journal = {arXiv preprint arXiv:2207.05221},
  year    = {2022}
}

@article{llava-med,
  title   = {L{L}ava-{M}ed: Training a {L}arge {L}anguage-and-{V}ision {A}ssistant for {B}iomedicine in {O}ne {D}ay},
  author  = {Li, Chunyuan and Wong, Cliff and Zhang, Sheng and Usuyama, Naoto and Liu, Haotian and Yang, Jianwei and Naumann, Tristan and Poon, Hoifung and Gao, Jianfeng},
  journal = {Advances in Neural Information Processing Systems},
  volume  = {36},
  year    = {2024}
}

@inproceedings{llavarad,
  title  = {Towards a clinically accessible radiology foundation model: open-access and lightweight, with automated evaluation},
  author = {Juan Manuel Zambrano Chaves and Shih-Cheng Huang and Yanbo Xu and Hanwen Xu and Naoto Usuyama and Sheng Zhang and Fei Wang and Yujia Xie and Mahmoud Khademi and Ziyi Yang and Hany Hassan Awadalla and Julia Gong and Houdong Hu and Jianwei Yang and Chunyuan Li and Jianfeng Gao and Yu Gu and Cliff Wong and Mu-Hsin Wei and Tristan Naumann and Muhao Chen and Matthew P. Lungren and Serena Yeung-Levy and Curtis P. Langlotz and Sheng Wang and Hoifung Poon},
  year   = {2024},
  url    = {https://api.semanticscholar.org/CorpusID:268379244}
}

@article{maira1,
  title   = {M{AIRA}-1: A specialised large multimodal model for radiology report generation},
  author  = {Hyland, Stephanie L and Bannur, Shruthi and Bouzid, Kenza and Castro, Daniel C and Ranjit, Mercy and Schwaighofer, Anton and P{\'e}rez-Garc{\'\i}a, Fernando and Salvatelli, Valentina and Srivastav, Shaury and Thieme, Anja and others},
  journal = {arXiv preprint arXiv:2311.13668},
  year    = {2023}
}

@article{medpalm,
  title     = {Towards {G}eneralist {B}iomedical {AI}},
  author    = {Tu, Tao and Azizi, Shekoofeh and Driess, Danny and Schaekermann, Mike and Amin, Mohamed and Chang, Pi-Chuan and Carroll, Andrew and Lau, Charles and Tanno, Ryutaro and Ktena, Ira and others},
  journal   = {NEJM AI},
  volume    = {1},
  number    = {3},
  pages     = {AIoa2300138},
  year      = {2024},
  publisher = {Massachusetts Medical Society}
}

@misc{mi2,
  title         = {MedImageInsight: An Open-Source Embedding Model for General Domain Medical Imaging},
  author        = {Noel C. F. Codella and Ying Jin and Shrey Jain and Yu Gu and Ho Hin Lee and Asma Ben Abacha and Alberto Santamaria-Pang and Will Guyman and Naiteek Sangani and Sheng Zhang and Hoifung Poon and Stephanie Hyland and Shruthi Bannur and Javier Alvarez-Valle and Xue Li and John Garrett and Alan McMillan and Gaurav Rajguru and Madhu Maddi and Nilesh Vijayrania and Rehaan Bhimai and Nick Mecklenburg and Rupal Jain and Daniel Holstein and Naveen Gaur and Vijay Aski and Jenq-Neng Hwang and Thomas Lin and Ivan Tarapov and Matthew Lungren and Mu Wei},
  year          = {2024},
  eprint        = {2410.06542},
  archiveprefix = {arXiv},
  primaryclass  = {eess.IV},
  url           = {https://arxiv.org/abs/2410.06542}
}

@inproceedings{rgrg,
  title     = {Interactive and explainable region-guided radiology report generation},
  author    = {Tanida, Tim and M{\"u}ller, Philip and Kaissis, Georgios and Rueckert, Daniel},
  booktitle = {Proceedings of the IEEE/CVF Conference on Computer Vision and Pattern Recognition},
  pages     = {7433--7442},
  year      = {2023}
}

@article{sellergren2025medgemma,
  title={Medgemma technical report},
  author={Sellergren, Andrew and Kazemzadeh, Sahar and Jaroensri, Tiam and Kiraly, Atilla and Traverse, Madeleine and Kohlberger, Timo and Xu, Shawn and Jamil, Fayaz and Hughes, C{\'\i}an and Lau, Charles and others},
  journal={arXiv preprint arXiv:2507.05201},
  year={2025}
}

@article{blasiok2023does,
  title={When does optimizing a proper loss yield calibration?},
  author={Blasiok, Jaroslaw and Gopalan, Parikshit and Hu, Lunjia and Nakkiran, Preetum},
  journal={Advances in Neural Information Processing Systems},
  volume={36},
  pages={72071--72095},
  year={2023}
}

@article{frohlich2024scoring,
  title={Scoring rules and calibration for imprecise probabilities},
  author={Fr{\"o}hlich, Christian and Williamson, Robert C},
  journal={arXiv preprint arXiv:2410.23001},
  year={2024}
}

@article{li2025conftuner,
  title={Conftuner: Training large language models to express their confidence verbally},
  author={Li, Yibo and Xiong, Miao and Wu, Jiaying and Hooi, Bryan},
  journal={arXiv preprint arXiv:2508.18847},
  year={2025}
}

@inproceedings{radrestruct,
  title={Rad-restruct: A novel vqa benchmark and method for structured radiology reporting},
  author={Pellegrini, Chantal and Keicher, Matthias and {\"O}zsoy, Ege and Navab, Nassir},
  booktitle={International Conference on Medical Image Computing and Computer-Assisted Intervention},
  pages={409--419},
  year={2023},
  organization={Springer}
}

@article{radialog,
  title={Radialog: A large vision-language model for radiology report generation and conversational assistance},
  author={Pellegrini, Chantal and {\"O}zsoy, Ege and Busam, Benjamin and Navab, Nassir and Keicher, Matthias},
  journal={arXiv preprint arXiv:2311.18681},
  year={2023}
}

@inproceedings{MedFlamingo23,
  author       = {Michael Moor and
                  Qian Huang and
                  Shirley Wu and
                  Michihiro Yasunaga and
                  Yash Dalmia and
                  Jure Leskovec and
                  Cyril Zakka and
                  Eduardo Pontes Reis and
                  Pranav Rajpurkar},
  editor       = {Stefan Hegselmann and
                  Antonio Parziale and
                  Divya Shanmugam and
                  Shengpu Tang and
                  Mercy Nyamewaa Asiedu and
                  Serina Chang and
                  Tom Hartvigsen and
                  Harvineet Singh},
  title        = {Med-Flamingo: a Multimodal Medical Few-shot Learner},
  booktitle    = {Machine Learning for Health, ML4H@NeurIPS 2023, 10 December 2023,
                  New Orleans, Louisiana, {USA}},
  series       = {Proceedings of Machine Learning Research},
  volume       = {225},
  pages        = {353--367},
  publisher    = {{PMLR}},
  year         = {2023},
  url          = {https://proceedings.mlr.press/v225/moor23a.html}
}

@inproceedings{davit,
  title={Da{V}i{T}: {D}ual attention vision transformers},
  author={Ding, Mingyu and Xiao, Bin and Codella, Noel and Luo, Ping and Wang, Jingdong and Yuan, Lu},
  booktitle={European conference on computer vision},
  pages={74--92},
  year={2022},
  organization={Springer}
}

@misc{unicl,
      title={Unified {C}ontrastive {L}earning in {I}mage-{T}ext-{L}abel {S}pace}, 
      author={Jianwei Yang and Chunyuan Li and Pengchuan Zhang and Bin Xiao and Ce Liu and Lu Yuan and Jianfeng Gao},
      year={2022},
      eprint={2204.03610},
      archivePrefix={arXiv},
      primaryClass={cs.CV}
}

@article{dathathri2019plug,
  title={Plug and play language models: A simple approach to controlled text generation},
  author={Dathathri, Sumanth and Madotto, Andrea and Lan, Janice and Hung, Jane and Frank, Eric and Molino, Piero and Yosinski, Jason and Liu, Rosanne},
  journal={arXiv preprint arXiv:1912.02164},
  year={2019}
}

@inproceedings{jindaug,
  title={DAug: Diffusion-based Channel Augmentation for Radiology Image Retrieval and Classification},
  author={Jin, Ying and Zhou, Zhuoran and Fang, Haoquan and Hwang, Jenq-Neng},
  booktitle={Advancements In Medical Foundation Models: Explainability, Robustness, Security, and Beyond},
  year={2024}
}

@inproceedings{jinqrad,
  title={QRad: Enhancing Radiology Report Generation by Captioning-to-VQA Reframing},
  author={Jin, Ying and Codella, Noel C and Xu, Yanbo and Gu, Yu and Wei, Mu and Fang, Haoquan and Lin, Thomas and Vozila, Paul and Hwang, Jenq-Neng},
  booktitle={The Second Workshop on GenAI for Health: Potential, Trust, and Policy Compliance}
}

@article{maira2,
  title={M{AIRA}-2: {G}rounded {R}adiology {R}eport {G}eneration},
  author={Bannur, Shruthi and Bouzid, Kenza and Castro, Daniel C and Schwaighofer, Anton and Bond-Taylor, Sam and Ilse, Maximilian and P{\'e}rez-Garc{\'\i}a, Fernando and Salvatelli, Valentina and Sharma, Harshita and Meissen, Felix and others},
  journal={arXiv preprint arXiv:2406.04449},
  year={2024}
}

@article{liu2026scaling,
  title={Scaling medical imaging report generation with multimodal reinforcement learning},
  author={Liu, Qianchu and Zhang, Sheng and Qin, Guanghui and Gu, Yu and Jin, Ying and Preston, Sam and Xu, Yanbo and Kiblawi, Sid and Yim, Wen-wai and Ossowski, Tim and others},
  journal={arXiv preprint arXiv:2601.17151},
  year={2026}
}

@inproceedings{yang2021fudge,
  title     = {{FUDGE}: Controlled Text Generation With Future Discriminators},
  author    = {Yang, Kevin and Klein, Dan},
  booktitle = {Proceedings of the 2021 Conference of the North American Chapter of the Association for Computational Linguistics: Human Language Technologies},
  pages     = {3511--3535},
  year      = {2021}
}

@inproceedings{xiong2024can,
  title     = {Can {LLMs} Express Their Uncertainty? An Empirical Evaluation of Confidence Elicitation in {LLMs}},
  author    = {Xiong, Miao and Hu, Zhiyuan and Lu, Xinyang and Li, Yifei and Fu, Jie and He, Junxian and Hooi, Bryan},
  booktitle = {International Conference on Learning Representations},
  year      = {2024}
}

@inproceedings{wang2023selfconsistency,
  title     = {Self-Consistency Improves Chain of Thought Reasoning in Language Models},
  author    = {Wang, Xuezhi and Wei, Jason and Schuurmans, Dale and Le, Quoc and Chi, Ed and Narang, Sharan and Chowdhery, Aakanksha and Zhou, Denny},
  booktitle = {International Conference on Learning Representations},
  year      = {2023}
}

@inproceedings{guo2017calibration,
  title     = {On Calibration of Modern Neural Networks},
  author    = {Guo, Chuan and Pleiss, Geoff and Sun, Yu and Weinberger, Kilian Q.},
  booktitle = {International Conference on Machine Learning},
  pages     = {1321--1330},
  year      = {2017}
}

@incollection{platt1999probabilistic,
  title     = {Probabilistic Outputs for Support Vector Machines and Comparisons to Regularized Likelihood Methods},
  author    = {Platt, John},
  booktitle = {Advances in Large Margin Classifiers},
  publisher = {MIT Press},
  year      = {1999}
}

@article{ho2022classifierfree,
  title   = {Classifier-Free Diffusion Guidance},
  author  = {Ho, Jonathan and Salimans, Tim},
  journal = {arXiv preprint arXiv:2207.12598},
  year    = {2022}
}

@article{wang2024trust,
  title   = {Trust It or Not: Confidence-Guided Automatic Radiology Report Generation},
  author  = {Wang, Yixin and Lin, Zihao and He, Jianzong and Zhang, Min and Wang, Ling and Li, Shuwen and Li, Jingyu and Zhou, S. Kevin and He, Zhiqiang},
  journal = {Neurocomputing},
  year    = {2024}
}

@article{zhang2020radiology_kg,
  title   = {When Radiology Report Generation Meets Knowledge Graph},
  author  = {Zhang, Yixiao and Wang, Xiaosong and Xu, Ziyue and Yu, Qihang and Yuille, Alan and Xu, Daguang},
  journal = {Proceedings of the AAAI Conference on Artificial Intelligence},
  year    = {2020}
}

@inproceedings{sun2025factaware,
  title     = {Fact-Aware Multimodal Retrieval Augmentation for Accurate Medical Radiology Report Generation},
  author    = {Sun, Liwen and Zhao, James Jialun and Han, Wenjing and Xiong, Chenyan},
  booktitle = {Proceedings of the 2025 Conference of the North American Chapter of the Association for Computational Linguistics},
  year      = {2025}
}

@article{lin2024trustworthy,
  title   = {Towards Trustworthy {LLMs}: a Review on Debiasing and Dehallucinating in Large Language Models},
  author  = {Lin, Zichao and Guan, Shuyan and Zhang, Wending and Zhang, Huiyan and Li, Yugang and Zhang, Huaping},
  journal = {Artificial Intelligence Review},
  year    = {2024}
}

@inproceedings{lau2018dataset,
  title     = {A Dataset of Clinically Generated Visual Questions and Answers about Radiology Images},
  author    = {Lau, Jason J. and Gayen, Soumya and Ben Abacha, Asma and Demner-Fushman, Dina},
  booktitle = {Scientific Data},
  year      = {2018}
}

@article{he2020pathvqa,
  title   = {{PathVQA}: 30000+ Questions for Medical Visual Question Answering},
  author  = {He, Xuehai and Zhang, Yichen and Mou, Luntian and Xing, Eric and Xie, Pengtao},
  journal = {arXiv preprint arXiv:2003.10286},
  year    = {2020}
}

@article{liu2021slake,
  title   = {{SLAKE}: A Semantically-Labeled Knowledge-Enhanced Dataset for Medical Visual Question Answering},
  author  = {Liu, Bo and Zhan, Li-Ming and Xu, Li and Ma, Lin and Yang, Yan and Wu, Xiao-Ming},
  journal = {arXiv preprint arXiv:2102.09542},
  year    = {2021}
}

@article{dsv4,
  title={Deepseek-v4: Towards highly efficient million-token context intelligence},
  author={Xu, Anyi and Lin, Bangcai and Xue, Bing and Wang, Bingxuan and Xu, Bingzheng and Wu, Bochao and Zhang, Bowei and Lin, Chaofan and Dong, Chen and Ling, Chenchen and others},
  journal={arXiv preprint arXiv:2606.19348},
  year={2026}
}
}

\clearpage
\appendix
\section{Sample Clinical User Interface}
\label{sec:sample_ui}

\begin{figure*}[hb!]
    \centering
    \includegraphics[width=\linewidth]{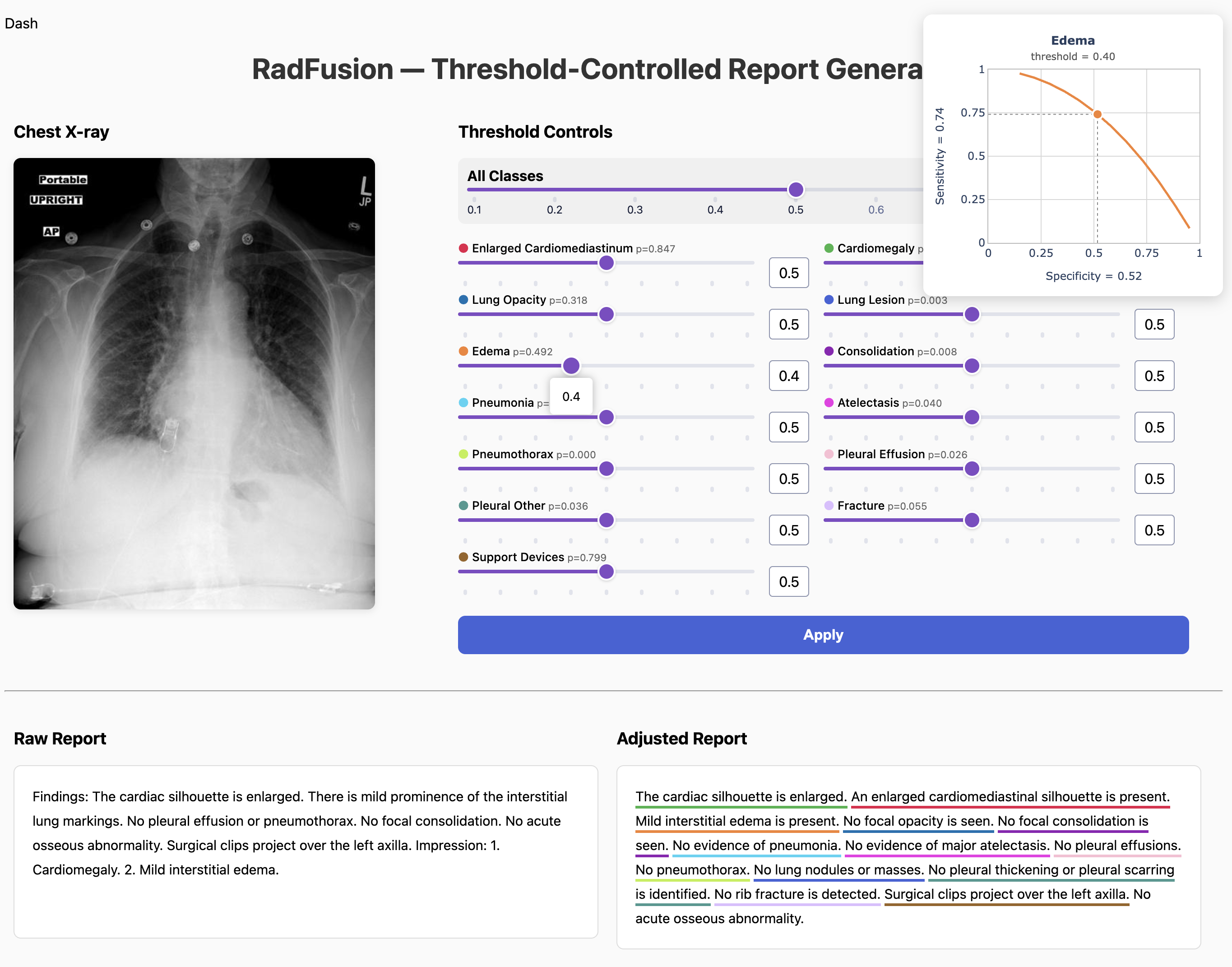}
    \caption{A sample clinical user interface for RadFusion. The radiograph and the automatically generated report are shown to the user, together with per-class threshold sliders annotated with the model's predicted confidence. Adjusting a slider highlights the corresponding operating point on the class's ROC curve, conveying the implied sensitivity--specificity trade-off. The adjusted report (bottom right) links each color-underlined text segment to its disease class, predicted confidence, and threshold on hover. Here, lowering the Edema threshold turns the class positive (third sentence, orange underlined)}
    \label{fig:sample_ui}
\end{figure*}

For additional context, we provide a sample user interface that demonstrates how RadFusion assists the daily radiology reporting process. As shown in \Cref{fig:sample_ui}, the radiograph and an automatically generated report are presented to the user. Alongside them, a set of threshold sliders for the disease classes is shown, each annotated with the model's predicted confidence. When the radiologist adjusts a slider, the corresponding operating point is highlighted on that class's ROC curve, giving the user immediate context on the implied sensitivity--specificity trade-off. The adjusted report is displayed on the bottom right, where hovering the mouse over each text segment (color-underlined) reveals the matching disease class, its predicted confidence, and the current threshold. In this example, the user tuned down the threshold for Edema, which turns the class positive (third sentence, orange underlined) in the adjusted report.

\section{Threshold-Control Evaluation on All Disease Classes}
\label{sec:full_roc}

\begin{figure}[t]
    \centering
    \includegraphics[width=0.93\linewidth]{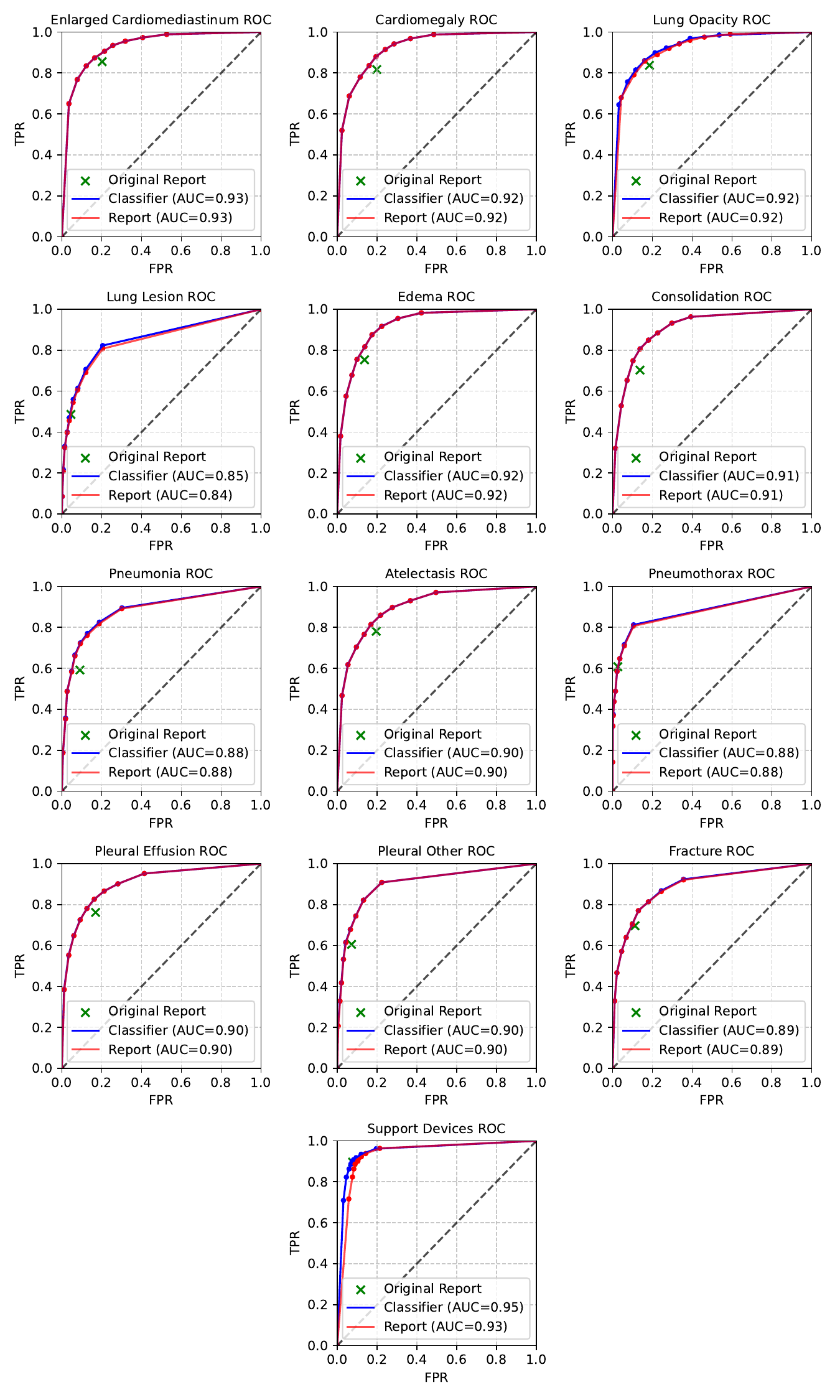}
    \caption{Threshold-controllable report generation on all 13 disease classes. High overlap between classifier ROC (blue) and report ROC (red) confirms conformance. The non-controlled report (green cross) falls below the curves. Ground-truth labels are obtained from the reference reports via VisualCheXbert.}
    \label{fig:threshold_ctrl_full_supp}
\end{figure}

\Cref{fig:threshold_ctrl_full_supp} presents the full threshold-control evaluation on all 13 disease classes, extending the four representative classes highlighted in the main text (\Cref{fig:threshold_ctrl_4_clases}). For each class, we compare the classifier ROC (blue), computed directly from the classifier's continuous confidence scores, against the ROC of the threshold-controlled reports (red), obtained by sweeping $\tau$ from 0.0 to 1.0 and converting each rewritten report back to class labels, as described in \Cref{sec:threshold_eval}. The green cross marks the single operating point of the non-controlled report. Across all classes, the report ROC closely tracks the classifier ROC, and the non-controlled report falls below the curves, confirming that the conformance and factual-correctness improvements reported in \Cref{sec:experiments} hold consistently across the full label set rather than only the highlighted classes.

\section{Comparison of Classifier Implementations}
\label{sec:roc_classifiers}

\begin{figure}[t]
    \centering
    \includegraphics[width=1\linewidth]{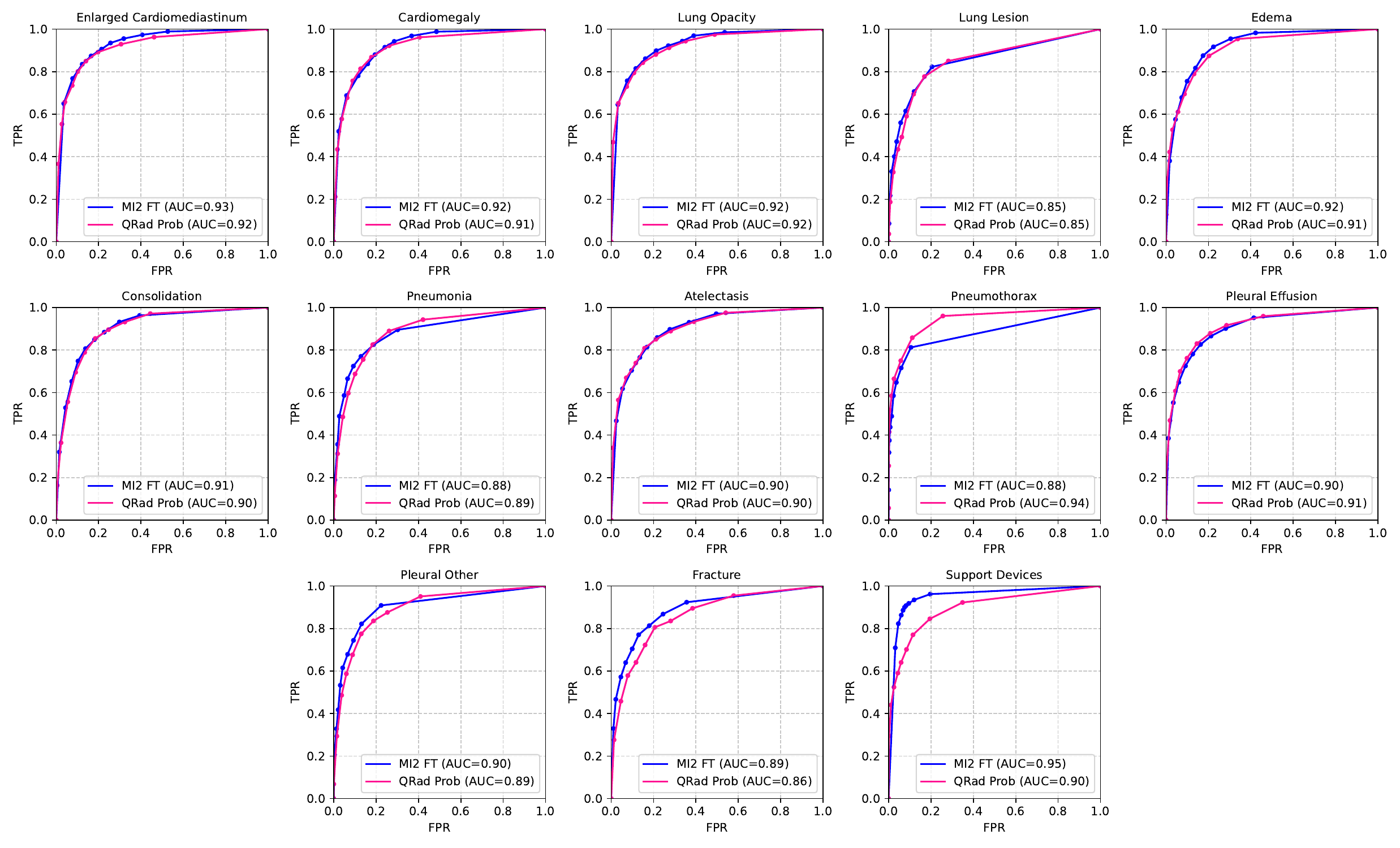}
    \caption{Per-class ROC curves of the fine-tuned MI2 classifier (MI2 FT) and a linear probe of \QRad's frozen encoder (\QRad Linear Prob) across the 13 disease classes. The two perform comparably across most classes, with \QRad Linear Prob stronger on hard classes such as Pneumothorax and MI2 FT stronger on visually obvious classes such as Support Devices.}
    \label{fig:roc_classifiers}
\end{figure}

\Cref{fig:roc_classifiers} shows the per-class ROC curves of the two strongest classifier implementations compared in \Cref{sec:classifier_abl}: the fine-tuned MI2 classifier (MI2 FT) and a linear probe of \QRad's frozen encoder (\QRad Linear Prob). These curves provide the full operating-point view underlying the summary AUC-ROC reported in \Cref{tab:roc_auc}. Consistent with the table, the two track each other closely across most classes, with \QRad Linear Prob pulling ahead on hard classes such as Pneumothorax and MI2 FT stronger on visually obvious classes such as Support Devices.

\section{Threshold-Control Evaluation with the Alternative Classifier}
\label{sec:full_roc_qrad}

\begin{figure}[t]
    \centering
    \includegraphics[width=1\linewidth]{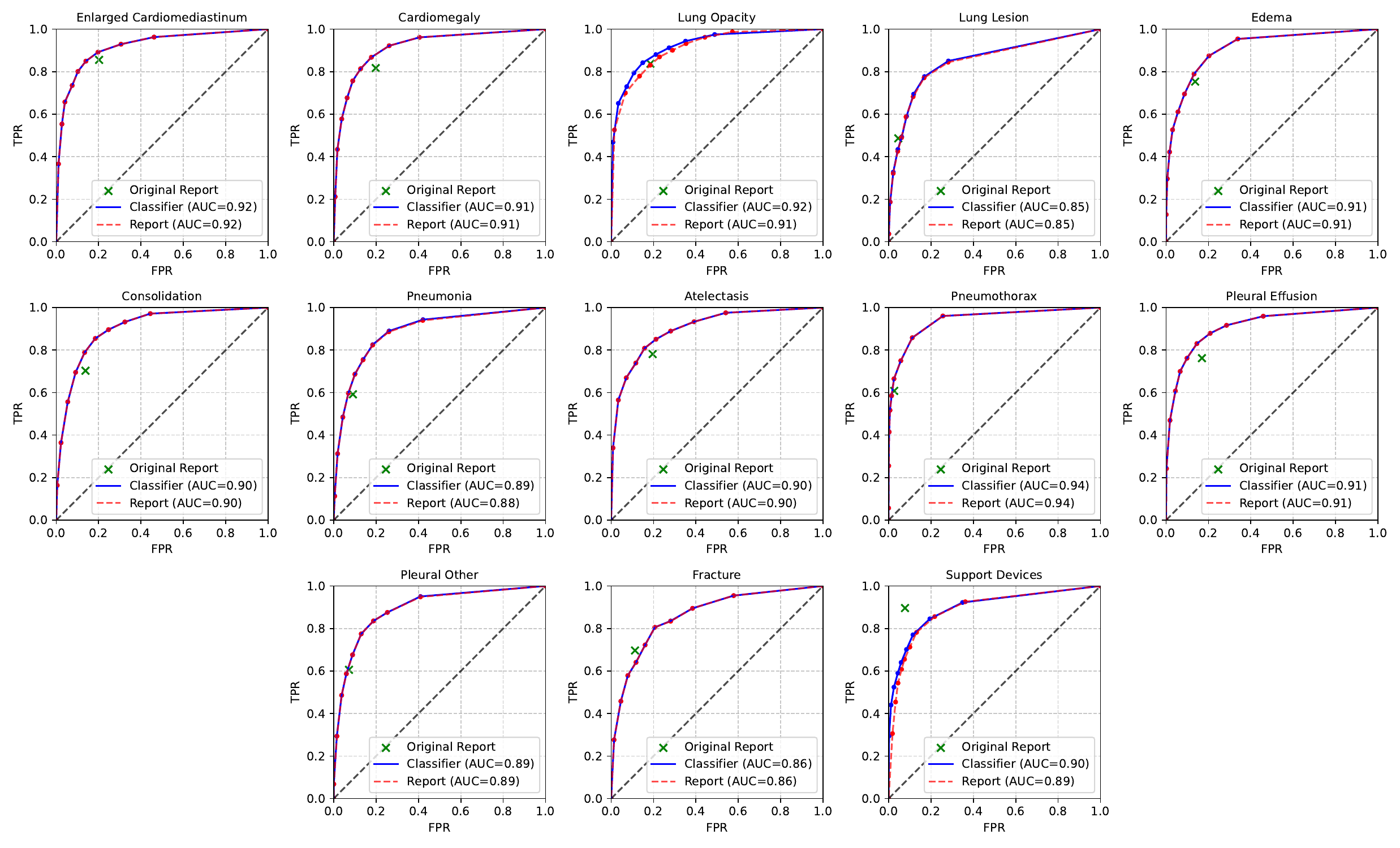}
    \caption{Threshold-control evaluation on all 13 disease classes using the \QRad Linear Prob classifier as the perception model. As with the default MI2 FT classifier, the report ROC (red) closely tracks the classifier ROC (blue), and the non-controlled report (green cross) falls below the curves. Ground-truth labels are obtained from the reference reports via VisualCheXbert.}
    \label{fig:full_roc_qrad}
\end{figure}

The threshold-control evaluation in \Cref{sec:full_roc} uses the default MI2 FT classifier as the perception model. Here we repeat it with the \QRad Linear Prob classifier (\Cref{sec:roc_classifiers}), an alternative implementation that shares its encoder with the report generator. \Cref{fig:full_roc_qrad} shows the per-class threshold-control evaluation across all 13 disease classes, with GPT-5 as the rewriter. As with the default classifier, the report ROC (red) closely tracks the classifier ROC (blue), and the non-controlled report (green cross) falls below the curves, confirming that RadFusion's threshold conformance holds across the underlying classifier.

\section{LLM Rewriting Instruction}
\label{sec:rewriting_prompt}

\Cref{fig:rewrite_prompt} shows the full instruction prompt provided to the LLM rewriter. The prompt consists of three parts: (1) a task description specifying the rewriting objective, (2) a template dictionary providing example positive and negative text segments for each of the 14 CheXbert disease classes, and (3) worked examples illustrating the rewriting rules for different cases. At inference, the placeholder \texttt{\{INPUT\}} is replaced with the original report and the threshold-adjusted classification results.

\begin{figure*}[t]
\footnotesize
\fontfamily{pcr}\selectfont
\begin{framed}
\raggedright
\fontfamily{phv}\selectfont
\textbf{System Instruction:}\par\vspace{0.4em}
\fontfamily{pcr}\selectfont
You are given an input chest x-ray radiology report and the classification result of 14 CheXbert classes, which includes both positive and negative diseases.\par\vspace{0.3em}
Modify the report so that it matches each class in the CheXbert classification results. Add text segments or sentences to the report if the class is not mentioned in the report, or change/remove sentences in the report so that it agrees with the classification.\par\vspace{0.3em}
Your modified report should be a valid Findings section of Chest X-ray radiology report, with information like the location, severity of medical findings based on the original report.\par\vspace{0.3em}
When overwriting the report with the positive and negative classes, consider using text segments from the following templates, and fit them into the report as appropriate.\par
\vspace{0.5em}
\fontfamily{phv}\selectfont\textbf{Template Dictionary} \fontfamily{pcr}\selectfont(abbreviated; full version includes all 14 classes):\par\vspace{0.3em}
\{\par
\hspace*{1em}"Cardiomegaly": \{\par
\hspace*{2em}"positive": ["The heart is moderately enlarged.",\par
\hspace*{4em}"Mild enlargement of the cardiac silhouette.", ...],\par
\hspace*{2em}"negative": ["The cardiac silhouette is normal.",\par
\hspace*{4em}"The heart is normal in size.", ...]\par
\hspace*{1em}\},\par
\hspace*{1em}"Pneumothorax": \{\par
\hspace*{2em}"positive": ["small apical pneumothorax", ...],\par
\hspace*{2em}"negative": ["no pneumothorax", ...]\par
\hspace*{1em}\}, ...\par
\}\par
\vspace{0.5em}
\fontfamily{phv}\selectfont\textbf{Rewriting Rules:}\par\vspace{0.3em}
\fontfamily{pcr}\selectfont
\textsl{Example 1:} If "Fracture" is a positive class:\par
\hspace*{1em}-- If the original report indicates Fracture is positive, do nothing.\par
\hspace*{1em}-- If the original report indicates Fracture is negative, replace that sentence with a positive text segment.\par
\hspace*{1em}-- If the original report doesn't mention Fracture, choose a positive text segment and add it to the report.\par
\vspace{0.3em}
\textsl{Example 2:} If "Support Devices" is a negative class:\par
\hspace*{1em}-- If the original report doesn't indicate the existence of any support device, do nothing.\par
\hspace*{1em}-- If the original report mentions any kind of support device exists, remove it.\par
\vspace{0.5em}
The input and output report are the Findings section of a radiology report. Make sure all positive classes are identifiable in the modified report. Return the modified version as JSON with "output" as the key.\par\vspace{0.3em}
Only return the JSON without any explanation. Double check and make sure you didn't forget any classes in the lists, and didn't make unnecessary changes:\par\vspace{0.3em}
\{INPUT\}
\end{framed}
\caption[LLM rewriting prompt for threshold-controllable report generation]{The instruction prompt for LLM-based report rewriting. The LLM receives the original report, threshold-adjusted classification results, and a template dictionary of example text segments for 14 CheXbert disease classes. The full template dictionary contains positive and negative text segments for all classes: Enlarged Cardiomediastinum, Cardiomegaly, Lung Opacity, Lung Lesion, Edema, Consolidation, Pneumonia, Atelectasis, Pneumothorax, Pleural Effusion, Pleural Other, Fracture, and Support Devices.}
\label{fig:rewrite_prompt}
\end{figure*}

\end{document}